# Embodied-Symbolic Contrastive Graph Self-Supervised Learning for Molecular Graphs

Daniel T. Chang (张遵)

*IBM (Retired)* dtchang43@gmail.com

**Abstract:** Dual embodied-symbolic concept representations are the foundation for deep learning and symbolic AI integration. We discuss the use of dual embodied-symbolic concept representations for molecular graph representation learning, specifically with exemplar-based contrastive self-supervised learning (SSL). The embodied representations are learned from molecular graphs, and the symbolic representations are learned from the corresponding Chemical knowledge graph (KG). We use the Chemical KG to enhance molecular graphs with symbolic (semantic) knowledge and generate their augmented molecular graphs.  We treat a molecular graph and its semantically augmented molecular graph as exemplars of the same semantic class, and use the pairs as positive pairs in exemplar-based contrastive SSL.

## 1 Introduction

Motivated by recent findings from cognitive neural science, we advocate the use of a *dual-level model for concept representations* [1]: the *embodied level* consists of concept-oriented feature representations, and the *symbolic level* consists of concept graphs (a.k.a. knowledge graphs). We further advocate the use of *dual embodied-symbolic concept representations* [3] for deep learning. That is, deep learning should learn from data not only *modality-specific embodied representations* such as image embeddings, graph embeddings, etc., but also the corresponding *amodal symbolic (semantic) representations* as knowledge graph embeddings. Dual embodied-symbolic concept representations are the foundation for *deep learning and symbolic AI integration*, which is an important direction for deep learning and AI since their integration reinforces each other's strength, compensates each other's weakness, and takes a major step toward human-level AI.

In this paper, we discuss the use of *dual embodied-symbolic concept representations* for *molecular graph representation learning*, specifically with *exemplar-based contrastive self-supervised learning (SSL)* [2]. The embodied representations are learned from *molecular graphs* which consist of *atoms* as nodes and (chemical) *bonds* as edges. The symbolic representations are learned from the corresponding *Chemical knowledge graph (KG)*. We use the Chemical KG to enhance molecular graphs with symbolic (semantic) knowledge and generate their *augmented molecular graphs*.  We treat a molecular graph and its semantically augmented molecular graph as *exemplars* of the same semantic class, and use the pairs as positive pairs in exemplar-based contrastive SSL.

*DGL* [4], *DGL-LifeSci* [5] and *DGL-KE* [6] are new open-source Python packages for deep graph learning, for deep graph learning in life sciences (particularly, molecular science), and for efficient generation of knowledge graph embeddings, respectively. They greatly facilitate the use of dual embodied-symbolic concept representations for molecular graph representation learning. We provide a brief review of each as background information.

*Graph SSL* [7] applies SSL methods to *graphs*. By training the *graph neural network (GNN)* model to solve well-designed pretext tasks from unlabeled data, Graph SSL helps the GNN model learn more generalized representations so it can achieve better performance on downstream tasks. Graphs have rich underlying *structural and attributive information* from which various pretext tasks can be designed. Furthermore, graphs are usually formed by domain-specific rules. Thus, *domain (symbolic) knowledge* can be incorporated into the design of pretext tasks and learning process. The learning strategies for Graph SSL can be divided into three categories: *Pre-training and Fine-tuning (P&F)*, *Joint Learning (JL)*, and (self-supervised) *Graph Representation Learning (GRL)*; whereas Graph SSL methods can be divided into three learning approaches: *contrastive, generative*, and *predictive*. We focus on the *GRL strategy* and the *contrastive learning approach*, in accordance with our focus on concept representation learning and our approach on contrastive SSL discussed in [1-2].

For *molecular graphs*, the associated domain (symbolic) knowledge is represented in *Chemical knowledge graph (KG)*. We focus on two fundamental kinds of Chemical KG: *Chemical Element KG* and *Functional Group KG*. They provide essential knowledge on physical chemical properties and chemical biological functions, respectively, of molecular graphs. The *Chemical Element KG* [10] provides *atom (node)-level* knowledge on physical chemical properties of molecular graphs. It is based on Periodic Table of Elements and describes *associations between atoms* that are not directly connected by bonds (edges) but related in fundamental physical chemical properties. The *Functional Group KG* [11-12] provides *functional group (subgraph)-level* knowledge on chemical biological functions of molecular graphs. *Functional Groups* are the *substituent* atoms or groups of atoms that are attached to specific molecules. They are responsible for the *chemical reactions* that the molecule they are attached to participate in.

The design of the *contrastive learning* approach to *Graph SSL* generally consists of three main modules: (1) graph augmentation, (2) pretext task, and (3) contrastive objective. *Embodied-symbolic contrastive Graph SSL* leverages *background (domain) knowledge* by using (1) *knowledge-enhanced* (i.e., *embodied-symbolic fused* [3]) graph augmentation and (2) *graph representation learning* as pretext task. *KCL (Knowledge-enhanced Contrastive Learning)* [10] is a framework that can be used for embodied-symbolic contrastive Graph SSL. It provides three modules: *knowledge-guided graph* *augmentation*, *knowledge-aware graph representation*, and *contrastive objective*. Knowledge-guided graph augmentation and encoding is core to the framework and domain specific, e.g., when applying the framework to molecular graphs. We discuss two fundamental kinds of *knowledge-guided molecular graph augmentation and encoding* based on, respectively, *Chemical Element KG* and *Functional Group KG*.

## 2 Background Information

### 2.1 DGL

*Deep Graph Library (DGL)* [4] is an open-source Python package designed specifically for *deep graph learning* (i.e., deep learning on graphs). It provides optimized implementations of well-known *graph neural network (GNN)* models, including graph convolutional network (GCN), relational GCN (R-GCN), graph attention network (GAT), and various message passing neural networks (MPNNs),. The different GNN models are unified by the *message passing paradigm*. When applying GNN models, there are two phases: a message passing phase and a readout phase. The *message passing phase* updates node representations simultaneously across the entire graph and consists of multiple iterations of message passing. The *readout phase* computes a representation for the entire graph. DGL graphs can be created from *NetworkX* graphs.

### 2.2 DGL-LifeSci

*DGL-LifeSci* [5] is an open-source Python toolkit for *deep graph learning* in *life sciences* (particularly, molecular science), based on DGL, PyTorch and RDKit. It allows GNN-based modeling of *molecular graph* data for prediction and generation tasks. DGL-LifeSci provides optimized modules for various stages of the molecular graph modeling pipeline. It provides high-quality and robust implementations of seven models for *molecular property prediction* (including GCN, MPNN, GAT, AttentiveFP and Weave), one model for *molecule generation*, and one model for *chemical reaction prediction*. DGL-LifeSci contains *four components*: (i) programming APIs to develop custom molecular graph modeling pipelines and models; (ii) a set of pretrained models that can either be fine-tuned or directly used; (iii) a set of built-in datasets for quick experimentation; and (iv) a set of ready-to-run scripts for training and prediction.

DGL-LifeSci provides built-in support for constructing *three kinds of graphs for molecules*: molecular graphs, distance-based graphs, and complete graphs. In all of these graphs, each *node* corresponds to an *atom* in a molecule. In a *molecular graph,* the *edges* correspond to *chemical bonds*. For graph featurization, it allows initializing various node and edge features from *atom and bond descriptors*.

### 2.3 DGL-KE

*DGL-KE* [6] is an open-source Python package for efficient generation of *knowledge graph embeddings (KGEs)*. It is implemented with Python on top of DGL along with a C++-based distributed key-value store. DGL-KE uses various optimizations to accelerate KGE generation on knowledge graphs with millions of nodes and billions of edges using multi-processing, multi-GPU, and distributed parallelism. These optimizations are designed to increase data locality, reduce communication overhead, overlap computations with memory accesses, and achieve high operation efficiency.

*KGE* [3] is a widely adopted approach to *knowledge graph (KG)* representation in which entities and relations in a KG are embedded in *low-dimensional continuous vector spaces*. Most KGE models create a vector for each entity and each relation. In DGL-KE, a *KG* is viewed as a set of statements (facts) having the form of *subject-predicate-object triples*, using the notation *(h, r, t) (head, relation, tail)* to identify a statement. It uses the subject-predicate-object triples present in the KG to generate the vector representations, (**h, r, t**), for (head, relation, tail). DGL-KE supports three *translational distance models* (TransE, TransR, RotatE) and three *semantic matching models* (RESCAL, DistMul, ComplEx).

## 3 Graph Self-Supervised Learning

Most of the work on deep graph learning has focused on *supervised learning*, in which GNN models are trained by specific downstream tasks with abundant labeled data. However, labeled data are often limited, expensive, and inaccessible. Furthermore, supervised learning methods, with specific downstream tasks, are prone to over-fitting, poor generalization, and weak robustness.

*Graph self-supervised learning (SSL)* [7] aims to overcome the above limitations by applying SSL methods to graphs. The primary goal of *SSL* is to learn transferable knowledge from abundant unlabeled data with well-designed pretext tasks and then generalize the learned knowledge to downstream tasks. By training the GNN model to solve well-designed pretext tasks from unlabeled data, Graph SSL helps the GNN model learn more generalized representations so it can achieve better performance on downstream tasks. Graphs have rich underlying *structural and attributive information*, from which various pretext tasks can be designed. Furthermore, graphs are usually formed by domain-specific rules, e.g., bonds (edges) between atoms (nodes) in molecular graphs are defined by chemical bond theory. Thus, *domain (symbolic) knowledge* can be incorporated a priori into the design of pretext tasks and learning process.

The learning strategies for Graph SSL can be divided into three categories [7]: *Pre-training and Fine-tuning (P&F)*, *Joint Learning (JL)*, and (self-supervised) *Graph Representation Learning (GRL)*. In all cases, *GNNs* are used as the backbone *encoder*. In the *P&F strategy*, the model is trained in a two-stage paradigm. At the pre-training stage, the encoder is pre-trained with the pretext tasks. At the fine-tuning stage, the pre-trained encoder is fine-tuned with a *prediction head* under the supervision of specific downstream tasks. In the *JL strategy*, the encoder is jointly trained with a prediction head under the supervision of pretext tasks and downstream tasks. Finally, in the *GRL strategy*, the model is also trained in a two-stage paradigm, with the first stage similar to Pre-training. However, at the second stage, the model is trained on the frozen pre-trained encoder with a projection head trained on downstream tasks only.

Graph SSL methods can be divided into three learning approaches [7]: *contrastive, generative*, and *predictive*:

- In the *contrastive learning* approach, multiple views are usually generated for each sample instance through various *graph augmentations*. Two views generated from the same instance are usually considered as a *positive pair*, while two views generated from different instances are considered as a *negative pair*. The primary goal of *contrastive learning* is to maximize the agreement of two jointly sampled positive pairs against the agreement of two independently sampled negative pairs. The design of the contrastive learning usually consists of three main modules: (1) graph augmentation, (2) pretext task, and (3) contrastive objective.
- In the *generative learning* approach, the prediction head is called the *graph decoder*, which is used to perform *graph reconstruction*. There are two categories of methods: (1) *graph autoencoder* that performs reconstruction in a once-for-all manner; (2) *graph autoregressive* that iteratively performs reconstruction. The *graph autoencoder* is trained to reconstruct certain parts of the input graph data, e.g., the *adjacency matrix* which stores the graph structure information and the relations between nodes. This is the approach taken by *Tiered Graph Autoencoders* [8-9].
- In the *predictive learning* approach, self-generated *informative labels* from the data are used as supervision to perform prediction tasks. There are four categories of labels: (1) Node Property labels, e.g., node degree, (2) Context-based labels, e.g., the shortest path length between nodes, (3) Self-Training labels, e.g., cluster from a previous learning stage, and (4) Domain Knowledge-based labels, e.g., functional group for molecular graphs.

In this paper, for Graph SSL, we focus on the *GRL strategy* and the *contrastive learning approach*, in accordance with our focus on concept representation learning and our approach on contrastive SSL discussed in [1-2]. In particular, for the contrastive learning approach, we focus on the *exemplar-based* approach [2] as it utilizes *semantically-valid graph augmentation* for the generation of positive pairs.

## 4 Chemical Knowledge Graphs

*Chemical knowledge graph embeddings* can be generated from *Chemical knowledge graphs (KGs)*. We focus on two fundamental kinds of Chemical KG: *Chemical Element KG* and *Functional Group KG*. They provide essential knowledge on physical chemical properties and chemical biological functions, respectively, of molecular graphs.

### 4.1 Chemical Element KG

The *Chemical Element KG* [10] provides *atom (node)-level* knowledge on physical chemical properties of molecular graphs. It is based on Periodic Table of Elements and describes *associations between atoms* that are not directly connected by bonds (edges) but related in fundamental physical chemical properties.

There are *108 elements* in the Chemical Element KG. Each element contains *17 types of physical chemical properties*, including family, metallicity, periodicity, state, weight, electronegativity, electron affinity, melting point, boiling point, ionization, radius, hardness, modulus, density, conductivity, heat, and abundance.

The statements (facts) in the Chemical Element KG are represented as *triples* in the form of *(property, relation, element)*, e.g., (Gas, isStateOf, Cl); For convenience of representation, continuous property values are discretized, resulting in a total of *107 property categories*. There are *17 relation types*, corresponding to the 17 types of physical chemical properties: isFamilyOf, isMetallicityOf, isPeriodOf, isStateOf, isWeightOf, isElectronegativityOf, isElectronAffinityOf, isMeltingPointOf, isBoilingPointOf, isIonizationOf, isRadiusOf, isHardnessOf, isModulusOf, isDensityOf, isConductivityOf, isHeatOf, and isAbundanceOf. In total, there are *1643 triples*.

### 4.2 Functional Group KG

The *Functional Group KG* provides *functional group (subgraph)-level* knowledge on chemical biological functions of molecular graphs. *Functional Groups* are the *substituent* atoms or groups of atoms that are attached to specific molecules. They are responsible for the *chemical reactions* that the molecule they are attached to participate in. Regardless of the molecule in which it is found, the same functional group will behave similarly and experience comparable chemical reactions. Thus, functional groups are the *moieties* which exhibit their own distinct features and properties independent of the molecule they are attached to. *Covalent bonding* links the atoms of functional groups and the functional group as a whole to the molecule.

The statements (facts) in the Functional Group KG [11-12] are represented as *relations*, specifically a collection of *logic programs* defining almost 100 relations for various *functional groups and rings* in a chemical compound (molecule with more than one type of chemical element), which for convenience are referred together as *moieties*. Functional groups are represented as *functional_group(CompoundID, Atoms, Length, Type)* and rings represented as *ring(CompoundID, RingID, Atoms, Length, Type)*. In addition, three higher level relations are defined to infer the presence of composite structures: *has_struc*(CompoundId, Atoms, Length, Struc), *fused*(CompoundId, Struc1, Atoms1, Struc2, Atoms2), and *connected*(CompoundId, Struc1, Atoms1, Struc2, Atoms2).

The Functional Group KG consists of *multiple hierarchies*. The hierarchy available in functional groups and rings is shown in Fig. 3 and Fig. 4 of [12], respectively. The most specific types of moieties (a total of 77) are also listed in Table 1 of [11].

## 5 Embodied-Symbolic Contrastive Graph SSL for Molecular Graphs

Leveraging *background knowledge* in Graph SSL is important in many applications. For example, leveraging *chemical knowledge* (e.g., chemical elements, functional groups) in molecular graph representation learning is crucial for chemical property prediction, molecular binding prediction, and drug discovery.

As discusses in Section 3 Graph Self-Supervised Learning, the design of the *contrastive learning* approach to *Graph SSL* usually consists of three main modules: (1) graph augmentation, (2) pretext task, and (3) contrastive objective. *Embodied-symbolic contrastive Graph SSL* leverages *background knowledge* by using (1) *knowledge-enhanced* (i.e., *embodied-symbolic fused* [3]) graph augmentation and (2) *graph representation learning* as pretext task.

*KCL (Knowledge-enhanced Contrastive Learning)* [10] is a framework that can be used for *embodied-symbolic contrastive Graph SSL*. It provides three modules:

- The *knowledge-guided graph augmentation* module leverages *Chemical KG* to guide the graph augmentation process. (Note that KCL specifically leverages Chemical Element KG. In this paper, we extend it to general Chemical KG.)

- The *knowledge-aware graph representation* module learns molecular representations. It adopts a commonly used *graph encoder* for the original molecular graphs and designs a *Knowledge-aware Message Passing Neural Network (KMPNN) encoder* to encode complex information in the augmented molecular graphs.
- The *contrastive objective* module trains the encoders to maximize the agreement between the two views (original and augmented) of molecular graphs.

Knowledge-guided graph augmentation and encoding is core to embodied-symbolic contrastive Graph SSL and domain specific, e.g., when applying the framework to molecular graphs. In the following, we discuss two fundamental kinds of *knowledge-guided molecular graph augmentation and encoding* based on, respectively, *Chemical Element KG* and *Functional Group KG*, Note that in both cases, we use the *Chemical KG* to enhance molecular graphs with symbolic (semantic) knowledge and generate their *augmented molecular graphs*. We treat a molecular graph and its semantically augmented molecular graph as *exemplars* of the same semantic class, and use the pairs as positive pairs in exemplar-based contrastive SSL.

## 5.1 Molecular Graph Augmentation and Encoding Based on Chemical Element KG

For molecular graph augmentation, *KCL* [10] extracts 1-hop neighbor *properties* of atoms (as chemical elements) in a molecule from *Chemical Element KG* and adds the corresponding *triples* as *new edges and nodes*. For example, it adds a node "Gas" and an edge from "Gas" to "Cl" to the original molecular graph based on the triple (Gas, isStateOf, Cl). (Note that the direction of each edge between the property and the atom is from the former to the latter, as shown in Figure 1 of [10].) This results in an *augmented molecular graph*, in which the original molecular structure is preserved, and neighborhood topologies for *atom-related properties* are introduced. The augmented molecular graph thus contains richer and more complex information, and is treated as a *positive sample* in contrastive learning.

In order to obtain the *initial features of properties and relations* in the augmented molecular graph, *KCL* adopts the KG embedding method, *RotatE*, to train *Chemical Element KG*. In this way, the initial features can capture the structural information of the *triples*.

The *augmented molecular graphs* are complex heterogeneous graph-structured data that fuses two types of information: the *embodied structural information* of molecular graphs, and the *symbolic domain knowledge* extracted from Chemical Element KG. *KCL* designs a *KMPNN encoder* to learn their graph-level representations. The key idea is to provide *two types of message passing* for different types of neighbors (*atoms* and *properties*), and assign them different attention according to their importance. It enables heterogeneous message passing with two MSG functions, where $MSG_a(.)$ is applied to neighbors representing atoms, and $MSG_p(.)$ is applied to neighbors representing properties.

## 5.2 Molecular Graph Augmentation and Encoding Based on Functional Group KG

A simple method to augment *molecular graphs* based on *Functional Group KG* is introduced in [11]: it inserts additional nodes with corresponding edges for each *functional group and ring*, jointly referred to as *moiety*, identified in a molecular graph.. (It makes use only of the most specific types of moieties.) This results in an *augmented molecular graph*, in which the original molecular structure is preserved, and neighborhood topologies for *atom-related moieties* as well as *moiety-to-moiety relations* (which are specific to the original molecular structure) are introduced. The augmented molecular graph thus contains richer and more complex information, and is treated as a *positive sample* in contrastive learning, same as the case based on Chemical Element FG.

To obtain the *augmented molecular graph*, the following *steps* are performed, as shown in Figure 1 of [11]:

1. Nodes for all *moieties* defined in the *Functional Group KG* are added to the molecular graph.
2. *Part-of edges* between the moiety nodes and their constituent atoms are added. Atoms can be part of multiple moieties.
3. The moieties are linked with an *edge* labeled as: a) *fused* when their constituent atom-level subgraphs share one or more nodes; or as b) *connected* when their constituent atom-level subgraphs do not have any node in common, but there exists an edge connecting nodes belonging to the two different moieties. Edges connecting any moiety to an *aliphatic chain* are not labeled as connected but as either c) *saturated* if the chain is saturated, or d) *unsaturated* otherwise.

The *augmented molecular graphs* are complex heterogeneous graph-structured data that fuses three types of information: the *embodied structural information* of molecular graphs, the *symbolic domain knowledge* extracted from Functional Group KG, and the *embodied-symbolic structural information* of moieties in molecular graphs. The structural information of moieties in molecular graphs, i.e., *edges linking moieties* as discussed in Step 3 above and shown in Figure 1 (c) of [11], is referred to as *moiety graphs*. An augmented molecular graph therefore consists of two graphs of different types: an *atom-bond graph* (the original molecular graph) and a *moiety graph*, as well as their interconnections, as shown in Figure 1 (d) of [11],

An *extended KMPNN encoder* is needed to learn the graph representations of *augmented molecular graphs*. It involves *four types of message passing* for different types of edges: *atom<->atom* for atom-bond graphs, *moiety<->moiety* for moiety graphs, and (*atom<-moiety, moiety<-atom*) for their interconnections; and assign them different attention according to their importance. It enables heterogeneous message passing with four MSG functions, where $MSG_a(.)$ is applied to atom<->atom edges, $MSG_m(.)$ is applied to moiety<->moiety edges, $MSG_{a<-m}(.)$ is applied to atom<-moiety edges, and $MSG_{m<-a}(.)$ is applied to moiety<-atom edges.

## 6 Conclusion

Dual embodied-symbolic concept representations are the foundation for deep learning and symbolic AI integration. They have been used successfully in computer vision for embodied-symbolic knowledge distillation for few-shot class incremental learning, and embodied-symbolic fused representation for image-text matching. In this paper, we show that they can be used successfully for molecular graph representation learning, specifically with exemplar-based contrastive self-supervised learning (SSL). The key is that we use the Chemical knowledge graph (KG), i.e., Chemical Element KG and/or Functional Group KG, to enhance molecular graphs with symbolic (semantic) knowledge and generate their augmented molecular graphs. We treat a molecular graph and its semantically augmented molecular graph as exemplars of the same semantic class, and use the pairs as positive pairs in exemplar-based contrastive SSL.

**Acknowledgement:** Thanks to my wife Hedy (郑期芳) for her support.